\documentclass[lettersize,journal]{IEEEtran}
\usepackage{amsmath,amsfonts}
\usepackage{algorithmic}
\usepackage{algorithm}
\usepackage{array}
\usepackage{textcomp}
\usepackage{stfloats}
\usepackage{amssymb}

\usepackage{verbatim}
\usepackage{graphicx}
\usepackage{cite}
\usepackage{mwe}  
\usepackage{subcaption}
\newcommand{\ra}[1]{\renewcommand{\arraystretch}{#1}}

\usepackage{multirow}
\usepackage{cite}
\usepackage{booktabs}
\usepackage{tabularx}
\usepackage{xcolor}
\usepackage{siunitx}

\definecolor{CATBlue}{HTML}{3E5CA4}
\definecolor{CATBlueDark}{HTML}{2F4F8F}
\definecolor{CATRed}{HTML}{C5232A}

\usepackage[
    colorlinks=true,
    linkcolor=CATBlueDark,
    citecolor=CATBlueDark,
    urlcolor=CATBlueDark
]{hyperref}
\usepackage[capitalise,nameinlink]{cleveref}

\begin{document}

\title{From Perception to Assistance: Open-Vocabulary Shared Autonomy for Robotic Manipulation}

\author{Murilo Vinicius da Silva$^{1}$, Ricardo V. Godoy$^{1,*}$, Juliano Negri$^{1}$, \\Gustavo J. G. Lahr$^{2}$, Ranulfo Bezerra$^{3}$, and Marcelo Becker$^{1}$ 
\thanks{This work was carried out with the support of Petrobras, using resources from the R\&D clause of the ANP, in partnership with the University of São Paulo and the intervening foundation FAFQ, under Cooperation Agreement No. 2023/00016-6 and 2023/00013-7.}
\thanks{$^{1}$Murilo Vinicius da Silva, Ricardo V. Godoy, Juliano Negri, and Marcelo Becker are with the Department of Mechanical Engineering,
        University of São Paulo, São Carlos, Brazil
        {\tt\small becker@sc.usp.br}}%
\thanks{$^{2}$Gustavo J. G. Lahr is with the Instituto Israelita de Ensino e Pesquisa Albert Einstein, Hospital Israelita Albert Einstein, São Paulo, Brazil}
\thanks{$^{3}$Ranulfo Bezerra is with the Graduate School of Information Sciences, Tohoku University, Sendai, Japan.}
\thanks{$^{*}$Corresponding author: {\tt\small ricardo.godoy@alumni.usp.br}}
}



\maketitle

\begin{abstract}
Teleoperating a robotic manipulator in industrial environments demands precision that camera-based interfaces alone struggle to deliver. The operator must align the end-effector with a target in clutter, under limited depth perception, and without colliding with the surrounding structures. This paper presents a shared-autonomy framework that assists the operator throughout this process. A single RGB-D camera captures the operator's arm motion and hand gestures without wearables, fiducials, or a calibration stage. The intended target is specified by a free-form text prompt, grounded by a vision-language model in the robot's gripper camera, and tracked across its onboard cameras by a promptable video-segmentation model, resulting in a grasp frame continuously separated from the obstacle map. Every commanded motion is executed by a GPU-accelerated model-predictive controller that enforces self- and environment-collision avoidance against an online volumetric reconstruction, while a potential field corrects the operator's reference toward the grounded target during the final approach. An autonomous mode can be gesture-triggered to complete the grasp on the same target without a separate perception pipeline. The framework is validated on a quadruped mobile manipulator. The interface achieves a positional RMSE of $59$\,mm relative to motion-capture ground truth, and the controller keeps the arm at least $18$\,cm from obstacles while the operator deliberately commands the arm into them by $6$\,cm. In an industrial valve manipulation and a pick-and-place task, the full framework succeeded in all trials, while ablating either the collision or the assistance module produced failures through complementary mechanisms, and autonomous execution succeeded in four of five trials per task. Project page: \url{https://perception-to-assistance.github.io}.
\end{abstract}


\section{Introduction} \label{sec:introduction}

Quadruped mobile manipulators are increasingly employed for industrial inspection and intervention~\cite{ramezani2020legged,miki2022learning, tranzatto2022cerberus}, which requires robots to navigate cluttered environments and interact with assets such as valves and tools. While the ability to traverse unstructured terrain and simultaneously execute dexterous tasks makes legged mobile manipulation uniquely suited to industrial settings, reliable manipulation in realistic workspaces remains challenging, as the system must couple locomotion stability, online perception in clutter, and contact-rich interaction within a unified framework~\cite{lopes2023review,sleiman2023versatile, 10325606,rigo2024hierarchical}.

\begin{figure}[!t]
    \centering
    \includegraphics[width=\linewidth]{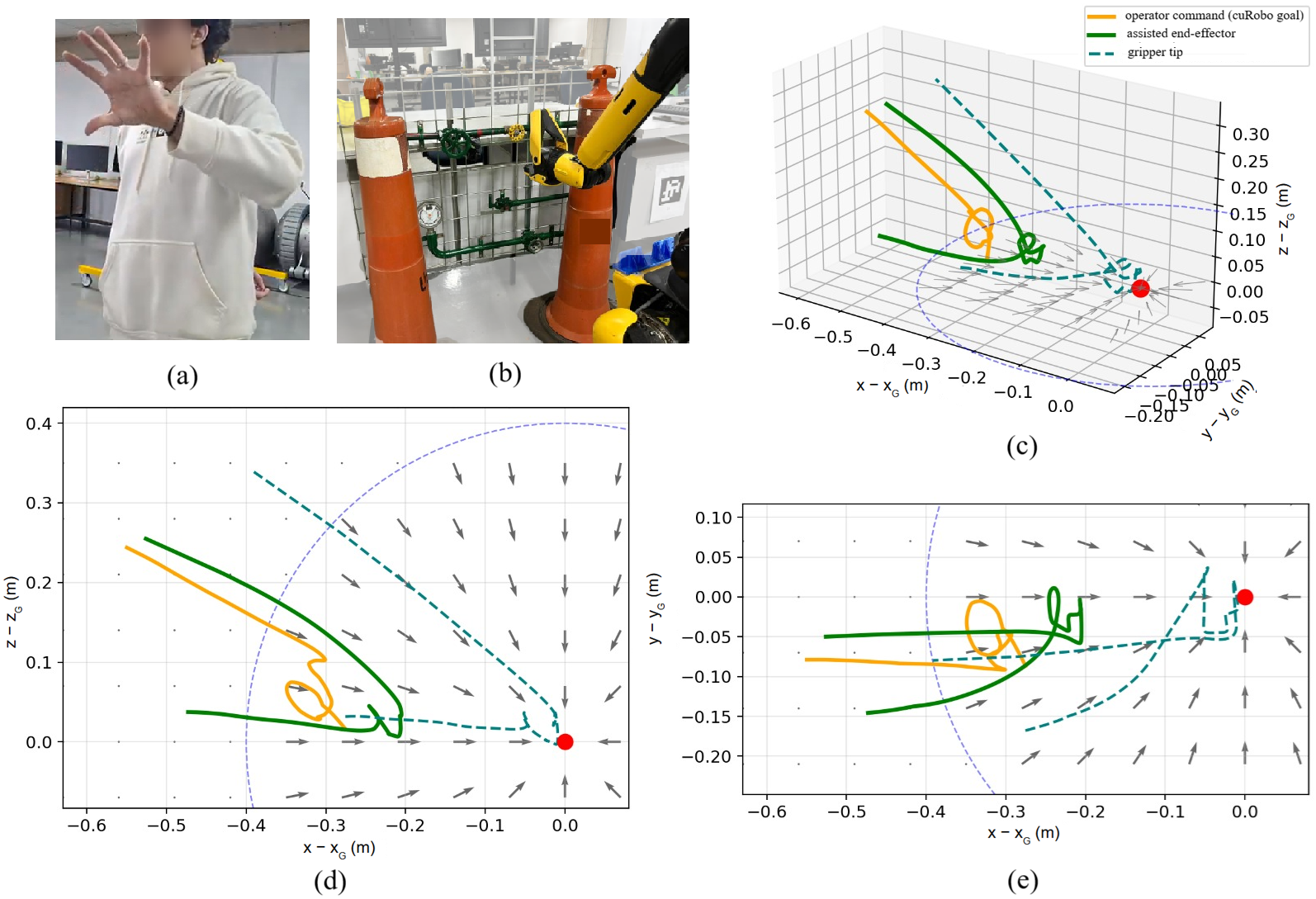}
    \caption{Shared-control valve manipulation. (a) The operator commands the robot via arm motion and hand gestures, captured by an RGB-D camera without wearables or calibration. (b) The robotic arm operates on an industrial panel in clutter. (c)--(e) Approach trajectories relative to the grasp frame $\{G\}$ (red), grounded from the prompt ``wheel valve''. Inside the $0.4$\,m activation radius (dashed circle), the potential field draws the assisted end-effector trajectory (green) toward the target while the operator command (orange) remains offset. The gripper tip (dashed) converges onto the target, and the loops correspond to the operator turning the valve after the grasp.}
\label{fig:cover}
\end{figure}

For manipulation in such environments, both pure teleoperation and full autonomy have important limitations. Teleoperation is flexible, but it places a high cognitive burden on the operator, who must simultaneously command a high-DoF arm, manage viewpoint and clearance, avoid collisions, and execute contact-rich motions under limited feedback~\cite{rubagotti2019semi,cruz2024analysis,7281253,zhou2024advancing}. Fully autonomous systems can execute scripted behaviors, but they often depend on reliable perception, precise extrinsics, known objects, and task parameters that are difficult to guarantee under domain shifts in unstructured environments~\cite{Krotkov2017DRC,kawatsuma2012emergency}. Shared-control addresses this gap by allowing the operator to specify intent while assistive modules handle local motion generation and safety through corrective interventions or arbitration, improving success rates and reducing collisions and user effort~\cite{guleccyuz2025enhancing, ozdamar2022shared,hagenow2021corrective, phung2024shared, 10342155}.

However, many shared-autonomy systems still rely on assumption-heavy target-selection pipelines (known objects, markers, predefined affordances) and low-rate collision representations, which limit safety and responsiveness in real industrial scenes~\cite{guleccyuz2025enhancing, ozdamar2022shared,hagenow2021corrective, 10341955}. For shared-control to remain usable during manipulation, the robot must perceive the environment onboard, continuously update a local collision representation, and react at interactive rates during execution, particularly for legged platforms whose advantage is operating outside fixed workcells. Recent onboard Truncated Signed Distance Field (TSDF)/Euclidean Signed Distance Function (ESDF) pipelines and learned distance-field representations make this level of continuous collision awareness increasingly practical~\cite{millane2024nvblox,ortiz2022isdf}.

A further challenge is semantic grounding and task parameterization. Predefined object classes, fiducials, and affordance templates do not generalize well across industrial sites. Open-vocabulary grounding with vision-language models (VLMs) enables target specification in natural language while handling visual variation~\cite{liu2024grounding}. Recent VLM-based models also show that semantic priors can support task-relevant reasoning and parameter extraction when coupled with executable control stacks~\cite{10.5555/3618408.3618748,zitkovich2023rt,huang2025a3vlm}, improve generalization in manufacturing and human-robot collaboration~\cite{fan2025vision}, and provide action-relevant structure such as articulation axes and constraints~\cite{huang2025a3vlm,buchanan2026online}. Nonetheless, the key unresolved problem is converting these semantic cues into continuous, safety-constrained assistance that remains valid during online replanning, viewpoint changes, and real manipulation.

In this paper we present a shared-autonomy system for mobile manipulation that remains robust in clutter and supports both assisted teleoperation and user-triggered autonomous execution. The robot continuously maintains a local 3D collision representation from onboard perception and performs online collision-aware motion generation to enable real-time obstacle avoidance during end-effector motion. To reduce operator workload during approach and alignment, we introduce a potential-field guidance layer that continuously corrects the operator's end-effector commands toward the intended target while preserving operator authority and enforcing collision constraints. We further generalize target specification by integrating an open-vocabulary perception module that grounds the operator's query in the scene and localizes a 3D grasp frame, which is used consistently for both shared-control guidance and autonomous task execution. Fig.~\ref{fig:cover} shows the framework in operation, in which the operator commands the arm through free motion and gestures, and the assisted end-effector is drawn toward the language-grounded target during a valve manipulation task. The autonomous execution module can be explicitly triggered via hand commands, enabling reliable task completion once intent is confirmed. We evaluate the complete pipeline on real industrial tasks, including wheel-valve operation and a drill pick-and-place, demonstrating safe, collision-aware manipulation and successful contact-rich execution in clutter.

The main contributions of this paper are:







\begin{itemize}
\item A shared-autonomy pipeline for mobile manipulation that couples calibration-free vision-based teleoperation, open-vocabulary target grounding, and gesture-triggered autonomous execution, converting a language query into a world-latched 3D grasp frame used to carry the operator from intent decoding to task completion.
\item Continuous collision-aware command tracking that integrates onboard volumetric mapping with sampling-based MPC, subjecting teleoperated and autonomous motion to the same constraints while the manipulation target is separated from the obstacle map online.
\item A potential-field assistance layer that corrects the operator's reference toward the grounded target within an activation region, with the operator's command remaining the dominant term at every control cycle.
\end{itemize}

\begin{figure*}[t]
  \centering
  \includegraphics[width=.92\textwidth]{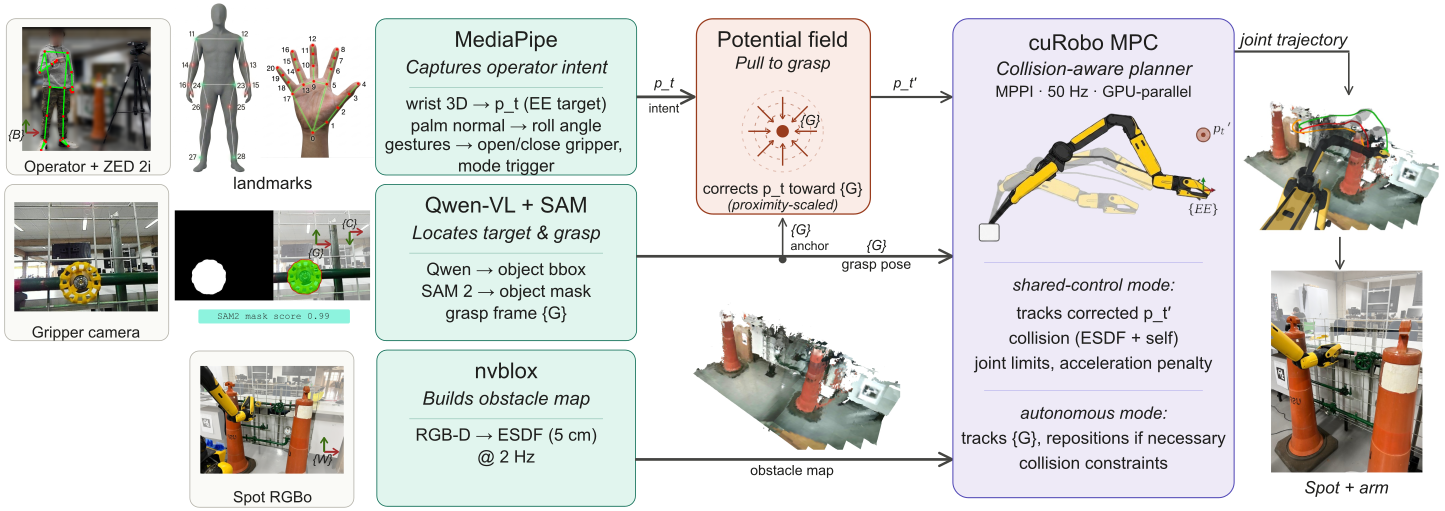}
  \caption{\textbf{System overview.} Three perception modules supply the controller. MediaPipe converts operator wrist motion, palm orientation, and hand gestures from a ZED 2i RGB-D feed into a teleoperation reference $\mathbf{p}_t$, gripper commands, and a mode trigger. The highlighted body landmarks (shoulders, hips, ankles) define the operator frame $\{B\}$, and the palm normal from the wrist and metacarpophalangeal landmarks drive the gripper roll. Qwen3-VL~\cite{bai2025qwen3} with SAM 2 grounds an open-vocabulary description against Spot's gripper camera, producing a grasp frame $\{G\}$ for objects such as the wheel valve illustrated. nvblox fuses Spot's stereo into a 5\,cm voxel TSDF and exposes an ESDF for distance queries. A proximity-scaled potential field corrects $\mathbf{p}_t$ toward $\{G\}$, producing the modulated reference $\mathbf{p}_t'$. cuRobo's MPPI-based MPC switches between two modes. Shared-control mode tracks $\mathbf{p}_t'$ under collision, joint-limit, and smoothness costs. Autonomous mode, engaged by the gesture trigger, tracks $\{G\}$ directly to complete the task. The output joint trajectory is sent to Spot's manipulator.}
  \label{fig:system_overview}
\end{figure*}


\section{Methods} \label{sec:methods}


The proposed shared-control framework, shown in Fig.~\ref{fig:system_overview}, decodes operator intent into safe arm motion via a modular perception-control pipeline. A vision-based teleoperation interface estimates a body-centered operator frame and maps wrist and hand motion to an end-effector position reference, gripper roll command, and discrete gripper actions. An onboard volumetric mapping module reconstructs a local 3D model of the workspace from the robot's stereo sensing for collision checking, and a GPU-accelerated reactive motion generator tracks the commanded reference at interactive rates while enforcing self- and environment-collision avoidance. On this backbone, an open-vocabulary grounding module and a potential-field assistance layer correct motion toward semantically grounded targets, and a user-triggered autonomous mode completes well-defined grasps. Each module is detailed below.

\subsection{Vision-Based Teleoperation Interface}
\label{sec:vision_mocap}

To teleoperate the robot arm without wearables or external motion-capture infrastructure, the proposed framework uses a ZED 2i RGB-D camera to acquire synchronized color and depth images. The pipeline can be adapted to any RGB-D sensor that provides aligned depth and known camera intrinsics. Rather than relying on fiducials or an explicit user calibration stage, the system continuously estimates a body-centered reference frame directly from the operator's pose.

\subsubsection{Body Tracking and Frame Estimation}

The RGB data is processed by MediaPipe Pose~\cite{mediapipe} to extract 33 skeletal landmarks, shown in Fig.~\ref{fig:system_overview}. We use the shoulders ($\mathbf{p}_{\mathrm{sh}}$), hips ($\mathbf{p}_{\mathrm{hp}}$), ankles ($\mathbf{p}_{\mathrm{ak}}$), and right wrist ($\mathbf{p}_{\mathrm{wr}}$). Each 2D landmark $(u,v)$ is deprojected into a 3D point $\mathbf{p}_{\mathrm{cam}}$ in the camera frame using the camera intrinsics $(f_x,f_y,c_x,c_y)$ and the median depth $d$ from an aligned $5 \times 5$ pixel neighborhood, as defined by Eq.~\ref{eq:deprojection}.

\begin{equation}
\mathbf{p}_{\mathrm{cam}} =
\begin{bmatrix}
(u-c_x)d/f_x \\
(v-c_y)d/f_y \\
d
\end{bmatrix}.
\label{eq:deprojection}
\end{equation}

A dynamic body-centered reference frame $\{B\}$ is then constructed from the operator's skeletal features. This approach requires no initial calibration step, as the frame is updated continuously from the detected pose, allowing free movement within the camera's field of view without degrading tracking quality. We define the centroids for the shoulders ($\mathbf{c}_{\mathrm{sh}}$), hips ($\mathbf{c}_{\mathrm{hp}}$), and ankles ($\mathbf{c}_{\mathrm{ak}}$) by averaging the left and right joint positions, as given by Eq.~\ref{eq:centroids}.

\begin{equation}\label{eq:centroids}
    \mathbf{c}_{i} = \frac{\mathbf{p}_{i_R} + \mathbf{p}_{i_L}}{2}, \quad \text{for } i \in \{\mathrm{sh}, \mathrm{hp}, \mathrm{ak}\}.
\end{equation}

To improve robustness under partial occlusion, the provisional vertical axis $\hat{\mathbf{u}}$ follows a hierarchical fallback, given by Eq.~\ref{eq:vertical_axis}. 

\begin{equation}
\hat{\mathbf{u}} =
\begin{cases}
\frac{\mathbf{c}_{\mathrm{sh}} - \mathbf{c}_{\mathrm{ak}}}{\|\mathbf{c}_{\mathrm{sh}} - \mathbf{c}_{\mathrm{ak}}\|} & \text{if ankles are visible,}\\[12pt]
\frac{\mathbf{c}_{\mathrm{sh}} - \mathbf{c}_{\mathrm{hp}}}{\|\mathbf{c}_{\mathrm{sh}} - \mathbf{c}_{\mathrm{hp}}\|} & \text{if ankles are occluded but hips are visible,} \\[12pt]
\hat{\mathbf{u}}_{\mathrm{cam}} & \text{if both hips and ankles are occluded.}
\end{cases}
\label{eq:vertical_axis}
\end{equation}

The lateral axis is always defined by the shoulder span, and the remaining axes are obtained through successive cross products, as given by Eq.~\ref{eq:body_frame}.

\begin{equation}
\hat{\mathbf{x}} =
\frac{\mathbf{p}_{\mathrm{sh}_R}-\mathbf{p}_{\mathrm{sh}_L}}
{\|\mathbf{p}_{\mathrm{sh}_R}-\mathbf{p}_{\mathrm{sh}_L}\|},
\qquad
\hat{\mathbf{z}} =
-\frac{\hat{\mathbf{x}} \times \hat{\mathbf{u}}}
{\|\hat{\mathbf{x}} \times \hat{\mathbf{u}}\|},
\qquad
\hat{\mathbf{y}} = \hat{\mathbf{x}} \times \hat{\mathbf{z}}.
\label{eq:body_frame}
\end{equation}

The origin is placed at the torso centroid and subsequently shifted to the right-shoulder projection so that the teleoperation frame is aligned with a virtual arm base analogous to the robot's shoulder.

\subsubsection{Landmark Filtering and Denoising}

Depth-based skeleton tracking often suffers from transient artifacts and outliers. To mitigate these in the reconstructed 3D landmarks, we employ a threshold-based rejection filter followed by temporal smoothing. Let $\Delta P_t$ denote the raw landmark spatial displacement at frame $t$, $v_t$ its instantaneous velocity, and $\Delta\theta$ the angular change of the frame axes. An update is accepted only if it satisfies spatial, kinematic, and angular constraints simultaneously, as given by Eq.~\ref{eq:update}.

\begin{equation}\label{eq:update}
C_t = \underbrace{\|\Delta P_t\| \leq 0.40}_{\text{dist. (m)}} \land \underbrace{v_t \leq 1.5}_{\text{vel. (m/s)}} \land \underbrace{\Delta\theta \leq 30^\circ}_{\text{ang.}}
\end{equation}

When $C_t$ fails, the landmark state $P^*_t$ holds its previous value. The accepted trajectory is then smoothed using an exponential moving average (EMA) filter, as given by Eq.~\ref{eq:filter}.

\begin{equation}\label{eq:filter}
\hat{P}_t = \alpha P^*_t + (1 - \alpha) \hat{P}_{t-1},
\end{equation}

where the smoothing factor $\alpha$ is empirically tuned within the range $[0.3, 0.5]$.

\subsubsection{End-Effector Command Generation}

The right-wrist landmark $\mathbf{p}_{\mathrm{wr}}^{\mathrm{cam}}$, obtained from Eq.~\ref{eq:deprojection}, is transformed into the body-centered frame $\{B\}$ using the rotation matrix $R_B~=~[\hat{\mathbf{x}} \ \hat{\mathbf{y}} \ \hat{\mathbf{z}}]$ from Eq.~\ref{eq:body_frame} and the origin $\mathbf{o}_B$. The resulting coordinate is scaled by a factor $s$, defined as the ratio between the robot arm's reachable workspace (approximately $1$\,m) and the operator's arm length, estimated online from the deprojected shoulder, elbow, and right-wrist landmarks as the sum of the upper-arm and forearm segment lengths, which are pose-invariant and therefore require no dedicated calibration step. The robot end-effector position reference is then computed as given by Eq.~\ref{eq:target_pose}.

\begin{equation}
\mathbf{p}_{\mathrm{target}}^{\mathrm{robot}} =
s\,R_B^\top \left(\mathbf{p}_{\mathrm{wr}}^{\mathrm{cam}} - \mathbf{o}_B\right)
+ \Delta \mathbf{p}_{\mathrm{sh}},
\label{eq:target_pose}
\end{equation}

where $\Delta \mathbf{p}_{\mathrm{sh}}$ is the constant shoulder offset of the robot arm with respect to the quadruped base.

\begin{figure}[t]
  \centering
  \includegraphics[width=.75\columnwidth]{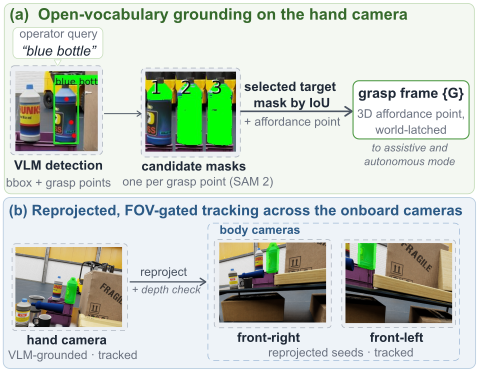}
  \caption{Open-vocabulary grounding and multi-camera tracking. (a) A free-form operator query is grounded by Qwen3-VL on the hand camera, which returns a bounding box and candidate grasp points. Each point is segmented by SAM~2, and the affordance point is selected by the best mask-to-box IoU, defining the world-latched grasp frame $\{G\}$. (b) The single hand camera detection is reprojected, gated on field of view and depth consistency, into the two body cameras.}
  \label{fig:grounding}
\end{figure}

\subsubsection{Hand Pose and Gripper Commands}

For hand orientation and grasp control, MediaPipe Hands~\cite{mediapipe} extracts 21 hand landmarks, as illustrated in Fig.~\ref{fig:system_overview}. The gripper roll reference $\phi$ is derived from the palm normal vector $\mathbf{n}$, computed from the wrist ($\mathbf{p}_0$), the index metacarpophalangeal joint ($\mathbf{p}_5$), and the little metacarpophalangeal joint ($\mathbf{p}_{17}$), as defined by Eq.~\ref{eq:roll}.

\begin{equation}
\mathbf{n} =
\frac{(\mathbf{p}_{17}-\mathbf{p}_0)\times(\mathbf{p}_5-\mathbf{p}_{17})}
{\|(\mathbf{p}_{17}-\mathbf{p}_0)\times(\mathbf{p}_5-\mathbf{p}_{17})\|},
\qquad
\phi = \mathrm{atan2}(n_y,n_x).
\label{eq:roll}
\end{equation}

A dedicated gesture-recognition module generates discrete gripper commands when \textit{Open Palm} and \textit{Closed Fist} gestures are detected. During these events, the roll command is briefly frozen to avoid transient orientation changes while the gripper is opening or closing. A separate finger-count gesture selects the operating mode. When the operator raises one finger, the shared-control teleoperation mode is engaged, while raising two fingers engages autonomous execution.

\subsection{Volumetric Mapping and Collision Representation}
\label{sec:perception}

The motion generator requires a continuously updated 3D model of the local workspace for collision avoidance. Since the quadruped operates outside fixed workcells, this model must be built from onboard sensing and remain robust to self-occlusions introduced by the arm during execution. We use \textit{nvblox}~\cite{millane2024nvblox}, a GPU-accelerated volumetric mapping framework, to fuse depth observations from the robot's onboard stereo cameras into a TSDF. The TSDF is built with a voxel size of $5$\,cm, chosen as a compromise between geometric fidelity and real-time performance during manipulation.

\paragraph{Local workspace integration}
The mapping volume is restricted to the region relevant for arm motion. The projective integration distance is limited to $1.5$\,m from the onboard cameras, removing distant background structures that are irrelevant for collision checking while preserving nearby scene geometry around the manipulation workspace, and reducing memory usage and stabilizing the local map during operation.

\paragraph{Interface to the motion generator}
The volumetric map is directly used by the motion generator via the ESDF, which is maintained and published at 2\,Hz. Inside the motion generation module, the incoming ESDF enables constant-time distance queries within the \textit{cuRobo} control loop for collision evaluation, so the controller always operates on the latest available local reconstruction even though the map refreshes at a lower rate than the control loop.

\subsection{Open-Vocabulary Object Grounding and Segmentation}
\label{sec:grounding}

The shared-control assistance requires a semantic target, i.e., the target object the operator intends to manipulate, localized in 3D and consistently separated from the surrounding scene. The target is obtained by grounding a natural-language description in the gripper-camera image with a VLM and re-projecting the resulting segmentation to the robot's other onboard camera feeds, as illustrated in Fig.~\ref{fig:grounding}.

\subsubsection{Language Grounding and Affordance Point}
The operator specifies the target with a free-form text prompt, for example ``wheel valve''. Qwen3-VL processes the gripper-camera image and returns, for the queried object, a bounding box, a confidence score, and a small set of candidate grasp (affordance) points, all in normalized image coordinates. Because the gripper camera rolls with the wrist, the image is first rotated to a gravity-upright orientation using the camera-to-world rotation obtained from the kinematic tree, and the detection is mapped back to the native image. Grounding is requested on demand and gated on view stability, low gripper-camera velocity and minimal frame latency, so that the VLM operates on a well-posed image.

\subsubsection{Multi-Camera Segmentation and Tracking}
The target is segmented continuously in the three onboard views, namely the gripper (\textit{hand}) camera and the two body cameras (\textit{frontleft} and \textit{frontright}), using a promptable video segmentation model, SAM\,2~\cite{sam2}, with one streaming predictor per camera. Each predictor maintains a temporal memory so the mask propagates frame to frame without re-querying the VLM. In deployment, a warm grounding call takes a median of $2.2$\,s, while each predictor sustains roughly $100$\,ms per inference for a mask rate of $2$--$2.5$\,Hz, matching the map update rate of Sec.~\ref{sec:perception}, with the $50$\,Hz controller tracking the world-latched frame in between.

On the hand camera, both the bounding box and the affordance points returned by the VLM are used. Each candidate grasp point is segmented and the affordance point is selected as the one whose mask agrees best, in intersection-over-union, with the bounding box, which yields a single point consistent with the detected object extent. To compensate for grounding latency, the selected affordance point is back-projected to a 3D point using the registered depth and camera intrinsics, then re-projected into the current frame via the kinematic tree before the streaming predictor initializes, so that the prompt lands on the object despite operator and robot motion.

The two body cameras are seeded by reprojecting the hand-tracked object position $\mathbf{p}_{\mathrm{obj}}$ into each secondary image plane (Fig.~\ref{fig:grounding}b), as given by Eq.~\ref{eq:reproj}.
\begin{equation}
u = f_x \frac{X}{Z} + c_x,\quad v = f_y \frac{Y}{Z} + c_y,\quad [X\ Y\ Z]^\top = R\,\mathbf{p}_{\mathrm{obj}} + \mathbf{t},
\label{eq:reproj}
\end{equation}
where $(R,\mathbf{t})$ is the world-to-camera transform at the frame timestamp. This provides a positive prompt and, on first acquisition, a bounding box sized from the object's measured extent at the expected depth $Z$. A depth-consistency check rejects a candidate mask whose measured depth disagrees with the reprojected expectation. Each camera publishes a per-frame binary mask stamped with the exact frame from which it was computed.

\subsubsection{Dynamic Object and Obstacle Separation}
The segmentation masks separate the manipulation target from the collision environment within the volumetric map of Sec.~\ref{sec:perception}. For every depth image, pixels inside the mask are fused into a \textit{dynamic} layer, while the remaining pixels build the \textit{static} TSDF from which the ESDF collision world is derived. The tracked object is therefore continuously excluded from the static obstacle representation. This separation is essential for collision-aware assistance, since a target fused into the static map would lead the motion generator to treat the very object the operator is reaching for as an obstacle and refuse to approach it. As a second, planning-side safeguard, the periodic ESDF query carries an axis-aligned bounding box that encloses the target, sized from the hand mask and its depth statistics, and is carved out of the map on each update, removing any residual single-frame leakage.

\subsubsection{Grasp Frame}
The 3D position of the affordance point defines the target grasp frame $\{G\}$, expressed in the world frame. To reject transient segmentation and depth noise under body motion, $\{G\}$ is world-latched, i.e., it holds a fixed position unless several consecutive measurements confirm that the object has genuinely moved, in which case it re-latches. This frame is the semantic target consumed by the potential-field assistance of Sec.~\ref{sec:pf}.

\subsubsection{Tracking Lifecycle}
The perception pipeline is governed by a lightweight state machine. The system is initialized on operator demand or when the target prompt changes: a grounding request runs the VLM, and the resulting seed places the per-camera predictors in a brief initialization window while they cold-start. Once a predictor locks onto the object, the system enters a tracking state in which the masks propagate through SAM\,2 memory and the grasp frame $\{G\}$ is updated. If the mask is lost, the target is no longer marked dynamic and would otherwise fuse into the static map, so the assistance freezes $\{G\}$ at its last world-latched position and the predictors attempt to re-acquire the object from their temporal memory, which recovers tracking without re-invoking the VLM once the object re-enters view. A new VLM grounding is requested only on an explicit operator trigger or a change of target, rather than automatically on every loss, so that transient occlusions during manipulation do not restart the pipeline.

\subsection{Collision-Aware Reactive Motion Generation}
\label{sec:rmg}

The control module converts the operator's Cartesian reference $\mathbf{p}_{\mathrm{target}}^{\mathrm{robot}}$ (Eq.~\ref{eq:target_pose}) and roll command $\phi$ (Eq.~\ref{eq:roll}) into feasible joint-space motion while continuously enforcing collision avoidance. We use \textit{cuRobo}~\cite{curobo_icra23}, which formulates reactive motion generation as a GPU-accelerated model predictive control (MPC) scheme. At each control cycle, the solver samples $N_p$ candidate control sequences using a Model Predictive Path Integral (MPPI) update. The optimization runs with an integration step of $dt=0.02$\,s, corresponding to a control frequency of $50$\,Hz, and each candidate predicts a trajectory over a horizon of $H$ steps. For every candidate trajectory, the GPU evaluates a multi-objective cost composed of: (i)~Cartesian tracking error with respect to the commanded end-effector reference, (ii)~joint-limit penalties, (iii)~smoothness penalties on the generated motion, and (iv)~collision costs associated with both self-collision and the environment. The final control action is obtained via standard MPPI cost-weighted averaging in a receding-horizon fashion, with lower-cost trajectories contributing more strongly to the executed command. In our implementation, $N_p = 400$ and $H = 24$.

\begin{figure}[t]
  \centering
  \includegraphics[width=.7\columnwidth]{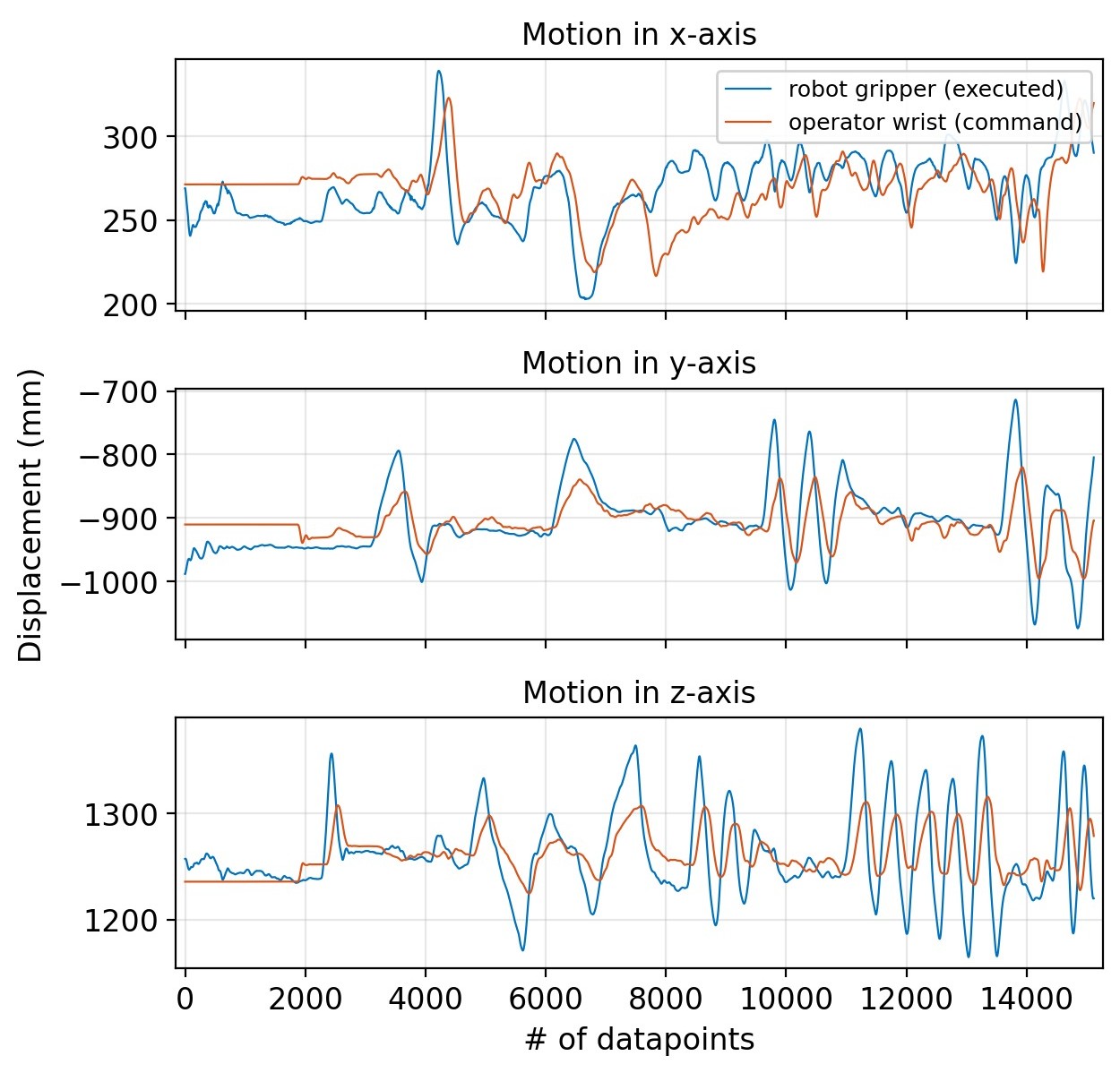}
  \caption{Teleoperation tracking accuracy. Operator wrist (command) and robot gripper (executed) displacement along each axis, measured by a six-camera OptiTrack system. The interface achieves a positional RMSE of $59$\,mm.}
  \label{fig:tracking_error}
\end{figure}

The robot geometry is represented by a set of collision spheres generated from the URDF using NVIDIA Isaac Sim's \textit{Lula Robot Description Editor}~\cite{isaac_lula_editor}. The end-effector objective is initialized relative to the current robot state, and inverse kinematics is solved locally around the current joint configuration. For environment collision checking, the controller queries the ESDF described in Sec.~\ref{sec:perception}. 


\subsection{Potential-Field Shared-Control Assistance}
\label{sec:pf}

The teleoperation interface of Sec.~\ref{sec:vision_mocap} lets the operator command the end-effector freely, but precise alignment with a target during the final approach is demanding under camera-based control. To reduce this effort while preserving operator authority, we introduce an attractive potential field that biases the operator's reference toward the grounded grasp frame $\{G\}$ (Sec.~\ref{sec:grounding}) as the end-effector nears it.

Let $\mathbf{p}_t\triangleq \mathbf{p}_{\mathrm{target}}^{\mathrm{robot}}$ denote the operator's commanded end-effector reference (Eq.~\ref{eq:target_pose}) and $\mathbf{p}_G$ the position of $\{G\}$, both in the robot body frame. The potential-field assistance engages only within an activation radius $d_a = 0.4$\,m of the target, following Eq.~\ref{eq:pf}.
\begin{equation}
\mathbf{p}_t' =
\begin{cases}
\mathbf{p}_t + k_{\mathrm{att}}\,(\mathbf{p}_G - \mathbf{p}_t), & \|\mathbf{p}_G - \mathbf{p}_t\| \le d_a,\\[4pt]
\mathbf{p}_t, & \text{otherwise,}
\end{cases}
\label{eq:pf}
\end{equation}
where $k_{\mathrm{att}}\in(0,1)$ sets the strength of the correction and $\mathbf{p}_t'$ is the modulated reference tracked by the motion generator. We set $k_{\mathrm{att}} = 0.2$. Only the attractive component is used, since repulsion is delegated to the collision costs of Sec.~\ref{sec:rmg}.

Since $k_{\mathrm{att}} < 1$, the operator's command remains the dominant term and authority is retained. The operator chooses when to approach and can always pull away, while the field only removes the fine residual error of the final centimeters. The distance is measured from the operator's reference rather than the current end-effector, so the assist engages based on where the operator is aiming. Because the correction is applied at every control cycle, the tracked reference is drawn smoothly toward $\{G\}$ as long as the operator holds the end-effector within $d_a$, producing a stable and progressively tightening approach, and since $\mathbf{p}_t'$ is tracked by the collision-aware generator of Sec.~\ref{sec:rmg}, the assisted motion remains subject to the same collision constraints.

\subsection{User-Triggered Autonomous Execution}
\label{sec:autonomous}

The shared-control assistance keeps the operator engaged throughout the approach. For well-defined or repetitive grasps, however, the operator can delegate the final approach and grasp to the robot once intent is confirmed. Following the gesture protocol of Sec.~\ref{sec:vision_mocap}, the operator raises two fingers to enter autonomous mode once the target has been grounded, and then initiates execution with a \textit{Closed Fist}, so autonomy engages only when the operator chooses to hand off. As in shared-control mode, an \textit{Open Palm} commands the gripper to release.

Autonomous execution reuses the pipeline rather than introducing a parallel one. The goal is the same grasp frame $\{G\}$ produced by the grounding module (Sec.~\ref{sec:grounding}), with no separate detector, fiducial, or hand-tuned target. The end-effector is brought to $\{G\}$ by the same collision-aware reactive motion generator of Sec.~\ref{sec:rmg}, the only difference being that the tracked reference is $\{G\}$ rather than the operator's modulated command $\mathbf{p}_t'$. The autonomous approach therefore inherits the identical self- and environment-collision guarantees enforced during teleoperation, and the target remains carved out of the static obstacle map (Sec.~\ref{sec:grounding}) so the controller can reach it without treating it as an obstacle.

Once the end-effector is aligned with $\{G\}$, the final grasp is delegated to Spot's onboard grasp service~\cite{BostonDynamicsSpotSDK}, seeded with the grounded target point. The perception and shared-control layers decide \emph{what} and \emph{where} to grasp, while the hardware-calibrated grasp primitive resolves the last-centimeter wrist alignment and closure, producing robust, repeatable grasps without reimplementing low-level grasp control. If the grounded target lies outside the current arm workspace, the grasp service can reposition the base to bring the object within reach, gated on operator confirmation and using Spot's onboard locomotion obstacle avoidance.

Throughout autonomous execution, the operator retains high-level authority and can intervene at any time, resuming shared-control teleoperation to complete contact-rich steps such as turning the grasped valve. 

\section{Experiments and Results} \label{sec:experiments}
\subsection{Experimental Setup}
\label{sec:setup}

All experiments were performed on a Boston Dynamics Spot quadruped robot equipped with the 6-DoF Spot Arm and its jaw gripper. The operator was tracked by a ZED 2i RGB-D camera placed in front of the operator workspace. Perception, grounding, mapping, and control ran on an offboard workstation with an NVIDIA RTX A2000 GPU ($12$\,GB) and an Intel Xeon CPU, communicating with the robot over Ethernet. The robot's base remained stationary during manipulation.

The framework was validated in three stages: the positional accuracy of the teleoperation interface against motion-capture ground truth, collision-aware shared control in a cluttered workspace, and the complete pipeline on two contact-rich manipulation tasks. Each manipulation task was executed under five conditions, pure teleoperation, teleoperation with only collision avoidance, teleoperation with only potential-field assistance, the full shared-control framework, and user-triggered autonomous execution, with five trials per condition performed by an operator. A trial is successful if the object is grasped and the task completed, and any contact with the surrounding structure ends the trial as a failure. The framework was further completed in all trials by a second user. 

A video demonstration of the experiments can be found at the following URL:
\begin{center}
\href{https://perception-to-assistance.github.io}{https://perception-to-assistance.github.io}
\end{center}

\subsection{Teleoperation Tracking Accuracy}
\label{sec:exp_accuracy}

To quantify how faithfully the interface transfers operator motion to the robot, reflective markers were attached to the operator's wrist and the Spot Arm's wrist, and both were tracked by an OptiTrack system comprising six PrimeX 41 cameras, which provides sub-millimeter 3D accuracy and which we employed as ground truth. The operator performed unscripted free motions across the workspace while the robot tracked the commanded reference through the full pipeline of Sec.~\ref{sec:vision_mocap} and Sec.~\ref{sec:rmg}. The positional error is computed between the operator's wrist trajectory, mapped into the robot workspace through the scale factor of Eq.~\ref{eq:target_pose}, and the robot end-effector trajectory, both measured by the OptiTrack system. The reported error therefore reflects the complete pipeline, including perception, filtering, scaling, and motion generation, rather than the tracking accuracy of the controller alone.

Fig.~\ref{fig:tracking_error} shows the wrist and gripper motion along each axis. The interface achieved a positional RMSE of $59$\,mm ($21$\,mm in $x$, $42$\,mm in $y$, and $35$\,mm in $z$), comparable to prior camera-based interfaces for legged manipulators~\cite{da_silva2025vision,zick2024teleoperation}, and the residual error is precisely what the assistance layer absorbs during the final approach. The gripper closely follows the commanded motion across all axes, including the fast direction reversals in the second half of the trial. The largest transient deviations coincide with these rapid reversals, where the smoothing filter and the finite tracking bandwidth of the arm introduce lag, while during slow, deliberate motion the executed trajectory stays close to the command. The error therefore behaves as lag rather than drift, and the interface remains accurate in the regime where teleoperation precision is paramount, the final approach to a target.

\begin{figure}[t]
  \centering
  \includegraphics[width=\columnwidth]{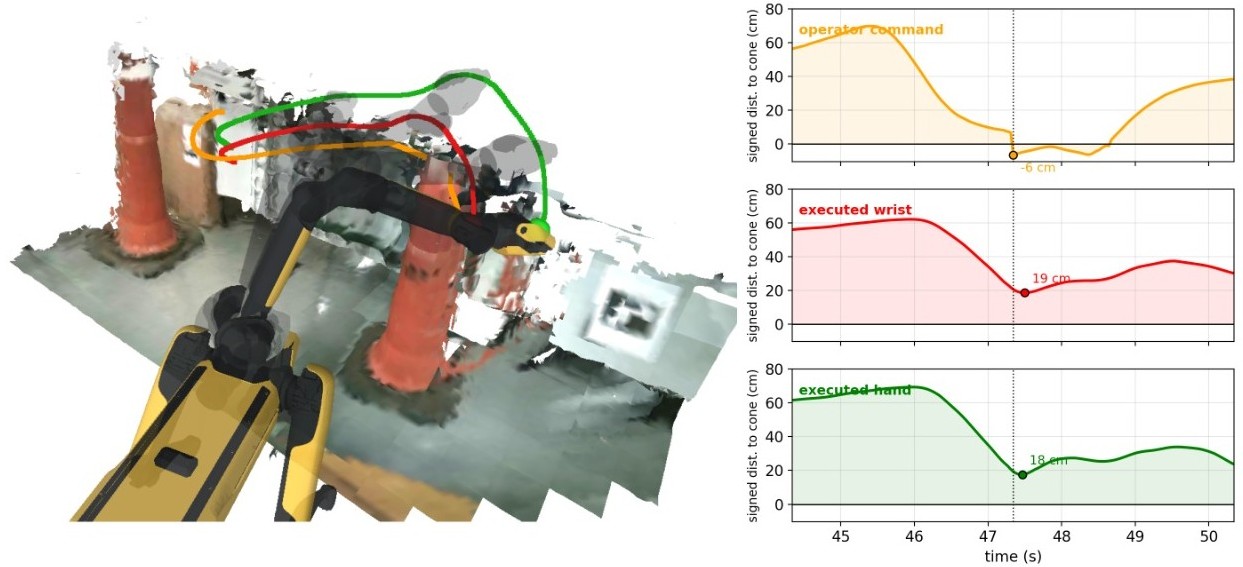}
  \caption{Collision-aware shared control. Left: reconstructed workspace with the commanded (orange), executed wrist (red), and executed hand (green) trajectories. Right: signed distance of each to the nearest cone over time. The operator deliberately drives the command $6$\,cm into the obstacle, while the executed wrist and hand never come closer than $19$\,cm and $18$\,cm.}
\label{fig:obstacle_avoidance}
\end{figure}

\subsection{Collision-Aware Shared Control}
\label{sec:exp_obstacles}

The second experiment demonstrates the collision avoidance of Sec.~\ref{sec:rmg} during teleoperation in clutter. The workspace, shown in Fig.~\ref{fig:cover}b, consists of a manipulation panel with industrial piping, valves, and a pressure gauge, flanked by traffic cones that constrain the free space around the arm. The operator commanded lateral end-effector motions across the workspace, deliberately sweeping toward and into the cones, while the signed distance of the commanded reference and of the executed arm links to the obstacles was recorded. Figure~\ref{fig:obstacle_avoidance} shows a representative sweep, with the reconstructed workspace, the commanded and executed trajectories, and the signed distance of each to the nearest cone over time. The commanded trajectory penetrates the cone surface, reaching $-6$\,cm at its deepest. The executed motion, in contrast, never approaches contact, where the wrist keeps a minimum distance of $19$\,cm and the hand $18$\,cm, as the MPPI collision costs correct the tracked path around the obstacle while continuing to follow the lateral component of the command. The deviation is confined to the vicinity of the cone, and tracking resumes as soon as the command returns to free space. The observed margin exceeds the collision-cost activation distance because distances are reported to the wrist and hand frames rather than to the collision-sphere surfaces, the ESDF voxelization slightly inflates the mapped surface, and the MPPI horizon penalizes approaching trajectories before the boundary is reached.

\subsection{Contact-Rich Manipulation Tasks}
\label{sec:exp_tasks}

The final experiment evaluates the complete framework on two tasks under the five conditions of Sec.~\ref{sec:setup}. In the assisted conditions the operator performs the approach and grasp, with the potential field active, while in the autonomous condition the operator delegates the approach and grasp to the robot (Sec.~\ref{sec:autonomous}). In all conditions the target is specified by the same free-form text prompt and grounded as in Sec.~\ref{sec:grounding}.

\subsubsection{Valve Manipulation}
The first task uses the industrial panel of Fig.~\ref{fig:cover}b. The operator grounds the target with the prompt ``wheel valve'', approaches, grasps the valve, and rotates it by a quarter turn. In the autonomous condition, the operator raises two fingers and confirms with a Closed Fist, after which the robot completes the approach and grasp on $\{G\}$, and in every condition the quarter-turn rotation is performed by the operator under shared control, consistent with Sec.~\ref{sec:autonomous}.
Fig.~\ref{fig:cover}c--e shows the assisted approach, plotting the operator's commanded reference, the assisted end-effector trajectory on which the field acts, and the gripper tip, all relative to $\{G\}$. Outside the $0.4$\,m activation radius the trajectories follow each other, since the field is inactive. Once the command enters the activation region, the assisted trajectory is drawn toward $\{G\}$, and the gripper tip converges onto the affordance point. The trailing loops correspond to the quarter-turn rotation performed by the operator after the grasp, executed under the same assistance. The commanded motion was amplified by the workspace scale factor $s$ of Eq.~\ref{eq:target_pose}, estimated online from the operator's arm length.

\subsubsection{Pick and Place}
The second task is a pick-and-place of a power drill resting on a box. The operator grounds the target with the prompt ``drill'', grasps it, teleoperates its transport, and releases it with an Open Palm.
This task complements the valve by requiring a free-object grasp, transport under load, and placement, rather than a fixed-fixture interaction.

\subsection{Ablation Study}

Table~\ref{tab:ablation} summarizes the outcomes of the ablation studies. Without either module, the limited depth perception of the camera interface makes the tasks largely infeasible, as the operator either contacts the surrounding structure or cannot align the gripper with the target. With collision avoidance alone, no contact occurs in any trial, but precise final alignment through the gripper camera remains difficult, and most trials fail at the grasp itself. With the potential field alone, the missed grasps largely disappear, but fast motions and fine corrections near the structure produce contacts that the assistive field cannot prevent. The two modules, therefore, suppress complementary failure modes, and with both active, all trials succeed. The autonomous condition failed once per task due to inaccurate depth estimates in the grasp service, which led to a misplaced final grasp pose, and the operator recovered by resuming shared control. In every trial, the grounding module localized the target from the free-form prompt, and the same pipeline handled a fixed, environment-mounted target and a free, movable object without retuning.

\begin{table}[h!]
\centering
\caption{Task success across the ablation conditions, five trials per condition. C denotes collision avoidance, and PF denotes the potential-field assistance.}
\ra{1.3}
\label{tab:ablation}
\begin{tabular}{lcc}
\hline
\textbf{Condition} & \textbf{Valve} & \textbf{Pick and place} \\
\hline
Teleoperation (no C, no PF) & 0/5 & 1/5 \\
Teleoperation + C & 2/5 & 2/5 \\
Teleoperation + PF & 3/5 & 4/5 \\
Full framework (C + PF) & \textbf{5/5} & \textbf{5/5} \\
Autonomous execution & 4/5 & 4/5 \\
\hline
\end{tabular}
\end{table}


\section{Conclusion} \label{sec:conclusion}

This paper presented a shared-autonomy framework for mobile manipulation that couples calibration-free vision-based teleoperation with open-vocabulary target grounding, continuous collision-aware motion generation, potential-field assistance, and a user-triggered autonomous execution mode, on a quadruped manipulator. The operator's motion is captured without wearables, fiducials, or a calibration stage. The intended target is specified by a free-form text prompt and tracked across the robot's onboard cameras, and every commanded motion, assisted or autonomous, is executed through the same collision-aware motion generator. The framework was validated on a Boston Dynamics Spot in three stages. The teleoperation interface achieved a positional RMSE of $59$\,mm against motion-capture ground truth. In a cluttered workspace, the motion generator kept the arm at least $18$\,cm from the obstacles while the operator deliberately commanded the reference $6$\,cm into them. In an industrial valve manipulation and a pick-and-place task, the full framework succeeded in all trials, while every ablated condition produced failures, through contact when the collision costs were removed and through missed grasps when the assistance was removed, showing that the two layers suppress complementary failure modes.

The current validation has limitations that define the next steps. The evaluation involved two operators and no external interface baseline, so a user study with more participants, comparisons against joystick and wearable-based teleoperation, and workload measures will be conducted in future work. Finally, the grounded grasp frame is a single affordance point, and richer geometric attributes, such as manipulation axes inferred from interaction rather than perception, would broaden the range of tasks the autonomous mode can complete without operator involvement.

\bibliographystyle{IEEEtran}
\bibliography{ref}

@inproceedings{curobo_icra23,
  author    = {Sundaralingam, Balakumar and Hari, Siva Kumar Sastry and
               Fishman, Adam and Garrett, Caelan and Van Wyk, Karl and Blukis, Valts and
               Millane, Alexander and Oleynikova, Helen and Handa, Ankur and
               Ramos, Fabio and Ratliff, Nathan and Fox, Dieter},
  booktitle = {2023 IEEE International Conference on Robotics and Automation (ICRA)},
  title     = {CuRobo: Parallelized Collision-Free Robot Motion Generation},
  year      = {2023},
  volume    = {},
  number    = {},
  pages     = {8112-8119},
  doi       = {10.1109/ICRA48891.2023.10160765}
}

@misc{mediapipe,
  author       = {{Google}},
  title        = {{MediaPipe}},
  howpublished = {\url{https://developers.google.com/mediapipe}},
  year         = {2024},
  note         = {Accessed: 2026-03-12}
}

@inproceedings{da_silva2025vision,
  author    = {da Silva, Murilo Vinicius and Carvalho, Matheus Hipolito and Negri, Juliano and Segreto, Thiago and Lahr, Gustavo J. G. and Godoy, Ricardo V. and Becker, Marcelo},
  title     = {A Vision-Based Shared-Control Teleoperation Scheme for Controlling the Robotic Arm of a Four-Legged Robot},
  booktitle = {2025 Latin American Robotics Symposium (LARS)},
  year      = {2025},
  doi       = {10.1109/LARS69345.2025.11272961}
}

@article{Krotkov2017DRC,
  author  = {Krotkov, Eric and Diftler, Myron and Ambrose, Robert and Deguet, Anton and Pires, Bruno and DeDonato, Matthew and Kazi, Zohaib and Kim, Won},
  title   = {The DARPA Robotics Challenge Finals: Results and Perspectives},
  journal = {Journal of Field Robotics},
  year    = {2017},
  doi     = {10.1002/rob.21734}
}

@article{rubagotti2019semi,
  author   = {Rubagotti, Matteo and Taunyazov, Tasbolat and Omarali, Bukeikhan and Shintemirov, Almas},
  journal  = {IEEE Robotics and Automation Letters},
  title    = {Semi-Autonomous Robot Teleoperation With Obstacle Avoidance via Model Predictive Control},
  year     = {2019},
  doi      = {10.1109/LRA.2019.2917707}
}

@article{miki2022learning,
  title={Learning robust perceptive locomotion for quadrupedal robots in the wild},
  author={Miki, Takahiro and Lee, Joonho and Hwangbo, Jemin and Wellhausen, Lorenz and Koltun, Vladlen and Hutter, Marco},
  journal={Science robotics},
  volume={7},
  number={62},
  pages={eabk2822},
  year={2022},
  publisher={American Association for the Advancement of Science}
}

@article{tranzatto2022cerberus,
  title={Cerberus in the darpa subterranean challenge},
  author={Tranzatto, Marco and Miki, Takahiro and Dharmadhikari, Mihir and Bernreiter, Lukas and Kulkarni, Mihir and Mascarich, Frank and Andersson, Olov and Khattak, Shehryar and Hutter, Marco and Siegwart, Roland and others},
  journal={Science Robotics},
  year={2022},
  publisher={American Association for the Advancement of Science}
}

@inproceedings{7281253,
  author    = {Jain, Siddarth and Farshchiansadegh, Ali and Broad, Alexander and Abdollahi, Farnaz and Mussa-Ivaldi, Ferdinando and Argall, Brenna},
  booktitle = {2015 IEEE International Conference on Rehabilitation Robotics (ICORR)},
  title     = {Assistive robotic manipulation through shared autonomy and a Body-Machine Interface},
  year      = {2015},
  volume    = {},
  number    = {},
  pages     = {526-531},
  doi       = {10.1109/ICORR.2015.7281253}
}

@article{kawatsuma2012emergency,
  title     = {Emergency response by robots to Fukushima-Daiichi accident: summary and lessons learned},
  author    = {Kawatsuma, Shinji and Fukushima, Mineo and Okada, Takashi},
  journal   = {Industrial Robot: An International Journal},
  year      = {2012},
  publisher = {Emerald Group Publishing Limited}
}

@inproceedings{ramezani2020legged,
  title={Legged robots for autonomous inspection and monitoring of offshore assets},
  author={Ramezani, Milad and Brandao, Martim and Casseau, Benoit and Havoutis, Ioannis and Fallon, Maurice},
  booktitle={Offshore Technology Conference},
  year={2020},
  organization={OTC}
}

@misc{BostonDynamicsSpotSDK,
  author       = {{Boston Dynamics}},
  title        = {{Spot SDK}: Software Development Kit for the Spot Robot},
  howpublished = {\url{https://github.com/boston-dynamics/spot-sdk}},
  year         = {2025},
  note         = {Accessed: 2025-05-31}
}

@article{sleiman2023versatile,
  title     = {Versatile multicontact planning and control for legged loco-manipulation},
  author    = {Sleiman, Jean-Pierre and Farshidian, Farbod and Hutter, Marco},
  journal   = {Science Robotics},
  volume    = {8},
  number    = {81},
  pages     = {eadg5014},
  year      = {2023},
  publisher = {American Association for the Advancement of Science}
}

@article{10325606,
  author   = {Jeon, Seunghun and Jung, Moonkyu and Choi, Suyoung and Kim, Beomjoon and Hwangbo, Jemin},
  journal  = {IEEE Robotics and Automation Letters},
  title    = {Learning Whole-Body Manipulation for Quadrupedal Robot},
  year     = {2024},
  volume   = {9},
  number   = {1},
  pages    = {699-706},
  doi      = {10.1109/LRA.2023.3335777}
}

@inproceedings{rigo2024hierarchical,
  title        = {Hierarchical optimization-based control for whole-body loco-manipulation of heavy objects},
  author       = {Rigo, Alberto and Hu, Muqun and Gupta, Satyandra K and Nguyen, Quan},
  booktitle    = {2024 IEEE International Conference on Robotics and Automation (ICRA)},
  pages        = {15322--15328},
  year         = {2024},
  organization = {IEEE}
}

@inproceedings{lopes2023review,
  title        = {A review on quadruped manipulators},
  author       = {Lopes, Maria S and Moreira, Ant{\'o}nio Paulo and Silva, Manuel F and Santos, Filipe},
  booktitle    = {EPIA Conference on Artificial Intelligence},
  pages        = {199--211},
  year         = {2023},
  organization = {Springer}
}

@article{zhou2024advancing,
  title     = {Advancing teleoperation for legged manipulation with wearable motion capture},
  author    = {Zhou, Chengxu and Wan, Yuhui and Peers, Christopher and Delfaki, Andromachi Maria and Kanoulas, Dimitrios},
  journal   = {Frontiers in Robotics and AI},
  volume    = {11},
  pages     = {1430842},
  year      = {2024},
  publisher = {Frontiers Media SA}
}

@inproceedings{10341955,
  author    = {Godoy, Ricardo V. and Guan, Bonnie and Dwivedi, Anany and Liarokapis, Minas},
  booktitle = {2023 IEEE/RSJ International Conference on Intelligent Robots and Systems (IROS)},
  title     = {An Affordances and Electromyography Based Telemanipulation Framework for Control of Robotic Arm-Hand Systems},
  year      = {2023},
  volume    = {},
  number    = {},
  pages     = {6998-7004},
  doi       = {10.1109/IROS55552.2023.10341955}
}

@inproceedings{ortiz2022isdf,
  title     = {{iSDF}: Real-time neural signed distance fields for robot perception},
  author    = {Ortiz, Joseph and Clegg, Alexander and Dong, Jing and Sucar, Edgar and Novotny, David and Zollhoefer, Michael and Mukadam, Mustafa},
  booktitle = {Robotics: Science and Systems (RSS)},
  year      = {2022}
}

@inproceedings{10.5555/3618408.3618748,
  author    = {Driess, Danny and Xia, Fei and Sajjadi, Mehdi S. M. and Lynch, Corey and Chowdhery, Aakanksha and Ichter, Brian and Wahid, Ayzaan and Tompson, Jonathan and Vuong, Quan and Yu, Tianhe and Huang, Wenlong and Chebotar, Yevgen and Sermanet, Pierre and Duckworth, Daniel and Levine, Sergey and Vanhoucke, Vincent and Hausman, Karol and Toussaint, Marc and Greff, Klaus and Zeng, Andy and Mordatch, Igor and Florence, Pete},
  title     = {PaLM-E: an embodied multimodal language model},
  year      = {2023},
  booktitle = {Proceedings of the 40th International Conference on Machine Learning},
  location  = {Honolulu, Hawaii, USA},
  series    = {ICML'23}
}

@inproceedings{10342155,
  author    = {Guan, Bonnie and Godoy, Ricardo V. and Sanches, Felipe and Dwivedi, Anany and Liarokapis, Minas},
  booktitle = {2023 IEEE/RSJ International Conference on Intelligent Robots and Systems (IROS)},
  title     = {On Semi-Autonomous Robotic Telemanipulation Employing Electromyography Based Motion Decoding and Potential Fields},
  year      = {2023},
  doi       = {10.1109/IROS55552.2023.10342155}
}

@inproceedings{huang2025a3vlm,
  title        = {A3VLM: Actionable Articulation-Aware Vision Language Model},
  author       = {Huang, Siyuan and Chang, Haonan and Liu, Yuhan and Zhu, Yimeng and Dong, Hao and Boularias, Abdeslam and Gao, Peng and Li, Hongsheng},
  booktitle    = {Conference on Robot Learning},
  pages        = {1675--1690},
  year         = {2025},
  organization = {PMLR}
}

@article{buchanan2026online,
  title={Online estimation and manipulation of articulated objects},
  author={Buchanan, Russell and R{\"o}fer, Adrian and Moura, Jo{\~a}o and Valada, Abhinav and Vijayakumar, Sethu},
  journal={Autonomous Robots},
  year={2026},
  publisher={Springer}
}

@article{fan2025vision,
  title     = {Vision-language model-based human-robot collaboration for smart manufacturing: A state-of-the-art survey},
  author    = {Fan, Junming and Yin, Yue and Wang, Tian and Dong, Wenhang and Zheng, Pai and Wang, Lihui},
  journal   = {Frontiers of Engineering Management},
  volume    = {12},
  number    = {1},
  pages     = {177--200},
  year      = {2025},
  publisher = {Springer}
}

@inproceedings{zitkovich2023rt,
  title        = {Rt-2: Vision-language-action models transfer web knowledge to robotic control},
  author       = {Zitkovich, Brianna and Yu, Tianhe and Xu, Sichun and Xu, Peng and Xiao, Ted and Xia, Fei and Wu, Jialin and Wohlhart, Paul and Welker, Stefan and Wahid, Ayzaan and others},
  booktitle    = {Conference on Robot Learning},
  pages        = {2165--2183},
  year         = {2023},
  organization = {PMLR}
}

@inproceedings{liu2024grounding,
  title        = {Grounding dino: Marrying dino with grounded pre-training for open-set object detection},
  author       = {Liu, Shilong and Zeng, Zhaoyang and Ren, Tianhe and Li, Feng and Zhang, Hao and Yang, Jie and Jiang, Qing and Li, Chunyuan and Yang, Jianwei and Su, Hang and others},
  booktitle    = {European conference on computer vision},
  pages        = {38--55},
  year         = {2024},
  organization = {Springer}
}

@inproceedings{millane2024nvblox,
  title        = {nvblox: Gpu-accelerated incremental signed distance field mapping},
  author       = {Millane, Alexander and Oleynikova, Helen and Wirbel, Emilie and Steiner, Remo and Ramasamy, Vikram and Tingdahl, David and Siegwart, Roland},
  booktitle    = {2024 IEEE International Conference on Robotics and Automation (ICRA)},
  pages        = {2698--2705},
  year         = {2024},
  organization = {IEEE}
}

@article{guleccyuz2025enhancing,
  title     = {Enhancing Shared Autonomy in Teleoperation under Network Delay: Transparency-and Confidence-Aware Arbitration},
  author    = {G{\"u}le{\c{c}}y{\"u}z, Ba{\c{s}}ak and Balachandran, Ribin and Panzirsch, Michael and Singh, Harsimran and Hulin, Thomas and Xu, Xiao and Steinbach, Eckehard},
  journal   = {IEEE Robotics and Automation Letters},
  year      = {2025},
  publisher = {IEEE}
}

@article{hagenow2021corrective,
  title     = {Corrective shared autonomy for addressing task variability},
  author    = {Hagenow, Michael and Senft, Emmanuel and Radwin, Robert and Gleicher, Michael and Mutlu, Bilge and Zinn, Michael},
  journal   = {IEEE robotics and automation letters},
  volume    = {6},
  number    = {2},
  pages     = {3720--3727},
  year      = {2021},
  publisher = {IEEE}
}

@article{phung2024shared,
  title     = {A shared autonomy system for precise and efficient remote underwater manipulation},
  author    = {Phung, Amy and Billings, Gideon and Daniele, Andrea F and Walter, Matthew R and Camilli, Richard},
  journal   = {IEEE Transactions on Robotics},
  year      = {2024},
  publisher = {IEEE}
}

@article{ozdamar2022shared,
  title     = {A shared autonomy reconfigurable control framework for telemanipulation of multi-arm systems},
  author    = {Ozdamar, Idil and Laghi, Marco and Grioli, Giorgio and Ajoudani, Arash and Catalano, Manuel G and Bicchi, Antonio},
  journal   = {IEEE Robotics and Automation Letters},
  volume    = {7},
  number    = {4},
  pages     = {9937--9944},
  year      = {2022},
  publisher = {IEEE}
}

@article{zick2024teleoperation,
  title     = {Teleoperation system for multiple robots with intuitive hand recognition interface},
  author    = {Zick, Lucas Alexandre and Martinelli, Dieisson and Schneider de Oliveira, Andr{\'e} and Cremer Kalempa, Vivian},
  journal   = {Scientific reports},
  year      = {2024},
  publisher = {Nature Publishing Group UK London}
}

@article{cruz2024analysis,
  title     = {Analysis of mr--vr tele-operation methods for legged-manipulator robots},
  author    = {Cruz Ulloa, Christyan and Dom{\'\i}nguez, David and del Cerro, Jaime and Barrientos, Antonio},
  journal   = {Virtual Reality},
  volume    = {28},
  number    = {3},
  pages     = {131},
  year      = {2024},
  publisher = {Springer}
}

@misc{isaac_lula_editor,
  author       = {{NVIDIA}},
  title        = {Isaac Sim {Lula} Robot Description Editor},
  howpublished = {\url{https://docs.isaacsim.omniverse.nvidia.com}},
  year         = {2024},
  note         = {Accessed: 2026-03-18}
}

@inproceedings{sam2,
  title={Sam 2: Segment anything in images and videos},
  author={Ravi, Nikhila and Gabeur, Valentin and Hu, Yuan-Ting and Hu, Ronghang and Ryali, Chaitanya and Ma, Tengyu and Khedr, Haitham and R{\"a}dle, Roman and Rolland, Chloe and Gustafson, Laura and others},
  booktitle={International Conference on Learning Representations},
  volume={2025},
  pages={28085--28128},
  year={2025}
}

@article{bai2025qwen3,
  title={Qwen3-vl technical report},
  author={Bai, Shuai and Cai, Yuxuan and Chen, Ruizhe and Chen, Keqin and Chen, Xionghui and Cheng, Zesen and Deng, Lianghao and Ding, Wei and Gao, Chang and Ge, Chunjiang and others},
  journal={arXiv preprint arXiv:2511.21631},
  year={2025}
}

\end{document}